\documentclass[lettersize,journal]{IEEEtran}
\usepackage{amsmath,amsfonts}
\usepackage{algorithmic}
\usepackage{algorithm}
\usepackage{array}
\usepackage[caption=false,font=normalsize,labelfont=sf,textfont=sf]{subfig}
\usepackage{textcomp}
\usepackage{stfloats}
\usepackage{url}
\usepackage{verbatim}
\usepackage{color}
\usepackage{bm}
\usepackage{graphicx}
\usepackage{multirow}
\usepackage{hyperref}
\usepackage{colortbl}
\usepackage{xcolor}
\usepackage{booktabs}
\usepackage{graphics} 
\usepackage{epsfig}
\usepackage{color}
\usepackage{svg}
\usepackage{orcidlink}
\usepackage{cite}

\begin{document}

\title{CoCoDiff: Diversifying Skeleton Action Features via Coarse-Fine Text-Co-Guided Latent Diffusion}

\author{Zhifu Zhao, Hanyang Hua, Jianan Li*, Shaoxin Wu, Fu Li, Yangtao Zhou, Yang Li
\thanks{This work was supported by the National Defense Basic Scientific
Research Program of China under Grant(JCKY2021413B005) and the National Natural Science Foundation of China (Grant No. 62302373, No.62202356, No. 62206210 and No. 62176201).} 
\thanks{(Corresponding author: Jianan Li.)}

\thanks{Jianan Li, Shaoxin Wu and Yangtao Zhou are with School of Computer Science and Technology, Xidian University, China (e-mail: lijianan@xidian.edu.cn).

Zhifu Zhao, Fu Li, and Yang Li are with School of Artificial Intelligence, Xidian University, China (e-mail: zfzhao@xidian.edu.cn).

Hanyang Hua is Guangzhou Institute of technology, Xidian University, China (e-mail: 22171214689@stu.xidian.edu.cn).

}}


\maketitle

\begin{abstract}
In action recognition tasks, feature diversity is essential for enhancing model generalization and performance. Existing methods typically promote feature diversity by expanding the training data in the sample space, which often leads to inefficiencies and semantic inconsistencies. To overcome these problems, we propose a novel \textbf{Co}arse-fine text \textbf{co}-guidance \textbf{Diff}usion model (\textbf{CoCoDiff}). 
CoCoDiff generates diverse yet semantically consistent features in the latent space by leveraging diffusion and multi-granularity textual guidance. 
Specifically, our approach feeds spatio‑temporal features extracted from skeleton sequences into a latent diffusion model to generate diverse action representations. Meanwhile, we introduce a coarse-fine text co‑guided strategy that leverages textual information from large language models (LLMs) to ensure semantic consistency between the generated features and the original inputs.
It is noted that 
CoCoDiff operates as a plug-and-play auxiliary module during training, incurring no additional inference cost. 
Extensive experiments demonstrate that CoCoDiff achieves SOTA performance on skeleton-based action recognition benchmarks, including NTU RGB+D, NTU RGB+D 120 and Kinetics-Skeleton.
\end{abstract}


\section{Introduction}
Human Action Recognition (HAR) is a task that classifies human actions in videos using video data as input. HAR is widely applied in fields such as human-computer interaction and virtual reality. In recent years, skeleton-based human action recognition~\cite{lee2023hierarchically,chen2021channel,yan2018spatial} has gained significant attention due to the development of depth sensors~\cite{zhang2012microsoft,keselman2017intel} and their robustness to complex backgrounds. However, when dealing with skeleton data, individual variations in height, body type, and movement habits introduce substantial variability into skeleton data, creating significant intra-class discrepancies.
\begin{figure}[t]
  \centering
   \includegraphics[width=0.9\linewidth]{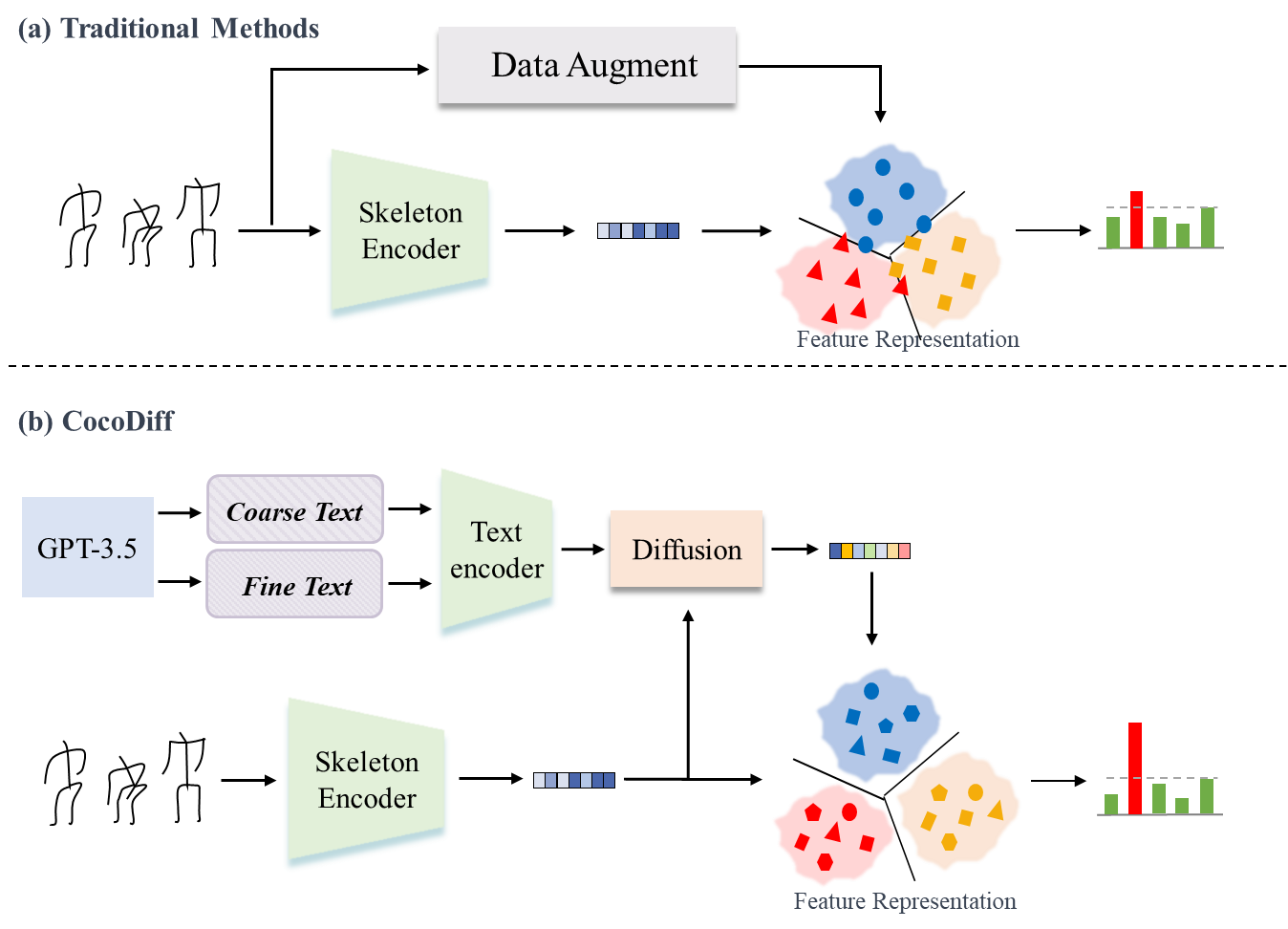}
   \caption{Comparison of the proposed Coarse-fine text co-guidance Diffusion model (CoCoDiff) framework with other traditional methods. The different colored geometric shapes in this figure represent action features of different categories.}
   \label{fig1}
\end{figure}

To improve generalization under large intra-class variations, it is necessary to provide the model with diverse yet semantically consistent action features. Previous approaches typically enhance feature diversity by augmenting the training data in the input sample space. They expand the training data by obtaining samples from different perspectives~\cite{pan2021view} and modalities~\cite{kocsar2023new}, applying data augmentation techniques~\cite{tu2018spatial}, or using generative models to generate new samples~\cite{xu2022generating}. 
However, these methods have the following problems: (1) \textbf{\textit{Computational Inefficiency:}} Extracting features from augmented samples often demands considerable computational resources making feature diversification inefficient in action recognition tasks, as the ultimate goal of action recognition tasks is to obtain discriminative features.
(2) \textbf{\textit{Semantic Inconsistency:}} Augmentation in the original sample space, particularly aggressive transformations (such as strong image transformations), may change the semantics of the samples, leading to a significant domain gap and the semantic inconsistency between synthetic and real samples.

To address the issues mentioned above, we propose generating diverse action features directly in the latent space using generative models. 
Unlike traditional sample-space augmentation, latent-space generation avoids costly data synthesis while preserving semantic fidelity. 
Diffusion model~\cite{ho2020denoising}, prominent in image synthesis tasks~\cite{rombach2021highresolution}, produce high-quality and diverse samples by progressively adding noise and learning reverse reconstruction iteratively. 
Spurred by their promising results, we propose integrating diffusion models into action recognition to generate diverse action features. 
However, directly using diffusion models to generate features may struggle to ensure semantic consistency, particularly in tasks characterized by large intra-class variation. For skeleton-based action recognition, the same action, such as drinking water, can exhibit significant differences across individuals in terms of posture, execution angle, and temporal dynamics. Without explicit semantic guidance, diffusion models are likely to capture, or even exaggerate, such intra-class variability during the generation process. Consequently, the synthesized features may suffer from semantic drift, leading to blurred class boundaries and degraded discriminative capacity.
To effectively mitigate such intra-class variability during feature generation, it is essential to introduce semantic-aware textual guidance that provides prior knowledge and helps anchor the generated features to the intended action semantics.

Therefore, we propose a \textbf{Co}arse-fine text \textbf{Co}-guidance \textbf{Diff}usion model (\textbf{CoCoDiff}) that addresses two critical limitations of traditional sample-space augmentation: computational inefficiency and semantic inconsistency. 
Guided by coarse- and fine-grained multi-scale text information, CoCoDiff fully capitalizes on the advantages of the diffusion model to effectively enhance feature diversity while preserving semantic consistency.
Figure \ref{fig1} shows the comparison between traditional methods and our proposed CoCoDiff model for enhancing feature diversity.
Specifically, we first use a skeleton encoder to extract skeletal features, which serve as input to the latent diffusion model for generating diverse action features. 
Then, the coarse-fine text co-guided strategy is performed in diffusion model to provide hierarchical textual supervision, anchoring the generative process and reducing ambiguity. Action descriptions are introduced as fine-grained textual guidance to provide contextual supervision. Simultaneously, action labels are incorporated as coarse-grained text to ensure semantic alignment. This dual-level textual supervision, implemented through contrastive learning, ensures that the generated features maintain both diversity and semantic consistency.
Notably, CoCoDiff is designed as a plug-and-play auxiliary training module and does not incur any additional cost during inference.
 
Our contributions can be summarized as follows:
\begin{itemize}
    \item \textbf{Latent Diffusion for Feature Generation:} We introduce a latent diffusion model for generating high-quality and diverse action features, effectively boosting model training and enhancing generalization. To the best of our knowledge, this is the first work to apply a latent diffusion approach to the action recognition task.
    \item \textbf{Coarse-to-Fine Text Co-Guidance Strategy:} We propose a novel coarse-to-fine text co-guidance strategy that jointly exploits coarse- and fine-grained textual information to guide feature generation, thereby reinforcing both semantic consistency and discriminative power.
    \item \textbf{State-of-the-Art Performance:} Comprehensive ablation experiments verify the effectiveness of CoCoDiff. Notably, our plug-and-play CoCoDiff achieves state-of-the-art performance on multiple skeleton-based action recognition benchmarks, including NTU RGB+D, NTU RGB+D 120, and Kinetics-Skeleton.
\end{itemize}

\section{Related Work}

\subsection{Skeleton-based Action Recognition}
In recent years, researchers have proposed various skeleton-based action recognition methods by designing deep models. Among them, methods based on Recurrent Neural Networks (RNNs)~\cite{Du_2015_CVPR,zhang2017geometric} typically model skeleton data as joint sequences under certain traversal methods to learn temporal dynamic relationships between frames. Inspired by the success of Convolutional Neural Networks (CNNs) in image tasks, CNN-based methods~\cite{zhang2019view, xu2021topology} have been utilized to model joints relations. Since human body joints can naturally be represented as graph nodes, and joint connections can be described using adjacency matrices, methods based on Graph Convolutional Networks (GCNs) have attracted significant attention. For instance, ST-GCN ~\cite{yan2018spatial} applies spatio-temporal GCNs to model human joint relationships in both spatial and temporal dimensions. With the recent popularity of Vision Transformers, Transformer-based methods (~\cite{plizzari2021spatial,shi2020decoupled,wang2023iip,wang20233mformer}) have also been applied to the field of skeleton-based action recognition. The performance of these methods heavily relies on the availability and quality of annotated skeleton data. Limited or noisy data can negatively impact on model performance. In this paper, the proposed method generates action features to alleviate the dependence of the model on training data.

\subsection{Latent Diffusion Models}
In recent years, the diffusion model has garnered widespread attention in the field of deep learning due to its outstanding performance. The original diffusion~\cite{ho2020denoising} model has a large number of parameters, leading to slow inference speed. Therefore, Rombach et al.~\cite{rombach2022high} propose the latent diffusion model, enabling the model to achieve faster training and inference on limited computational resources. Subsequently, AudioLDM~\cite{liu2023audioldm} is the first one to apply latent diffusion models in the field of text audio generation. Currently, latent diffusion models are mainly applied to low-level tasks such as generation. There is relatively less work applying diffusion models to high-level tasks. For example, DiffusionDet~\cite{chen2023diffusiondet} describes object detection as a denoising diffusion process from a noisy box to a target box. Recently, Wang et al.~\cite{wang2023diffusion} propose a new diffusion recommendation model to learn the generation process in a denoising manner. Despite the significant achievements of diffusion model in various fields, they are still seldom used in action recognition task. For instance, Jiang et al.~\cite{jiang2023spatial} employe diffusion models to generate new skeleton data, augmenting the original dataset. In this paper, the proposed method utilizes high-level feature representations extracted by GCN as inputs to the latent diffusion model, rather than using the raw skeleton data as input. This approach reduces computational costs while generating diverse action features.

\subsection{Conditional Generation Model}
Early generative work focuses on modeling unconditional, single-category data distributions, such as handwritten digits, certain animal species, and faces~\cite{deng2012mnist,choi2020stargan,karras2019style}. While unconditional models can rapidly achieve realistic results in single-category data, research indicates that unconditional generative models often suffer from mode collapse when extending to multi-category or diverse real image distributions~\cite{casanova2021instance}. To address the issue of mode collapse, conditional generative models are introduced. Various types of data have been employed as conditions for generative models, including category labels, image instances, and even networks~\cite{mirza2014conditional,casanova2021instance}. Meanwhile, CLIP~\cite{radford2021learning}, a large-scale pretrained image-text contrastive model, offers highly diverse text-image priors and has been found to be effective as a condition for generative models~\cite{crowson2022vqgan}. Recently, models like DALL-E 2~\cite{ramesh2022hierarchical} and Stable Diffusion~\cite{rombach2022high} can generate high-quality images solely based on free-form text, inheriting the diversity obtained from billions of real images collected from the internet. Currently, most conditional generative diffusion models incorporate various modalities of conditional information only during the diffusion process. For classification tasks, the model may lack sufficient guidance information. Thus, we introduce additional task-orientated information to the model, enhancing the semantic consistency between the generated results and input data.

\begin{figure*}[t]
\centering
\includegraphics[width=0.9\linewidth]{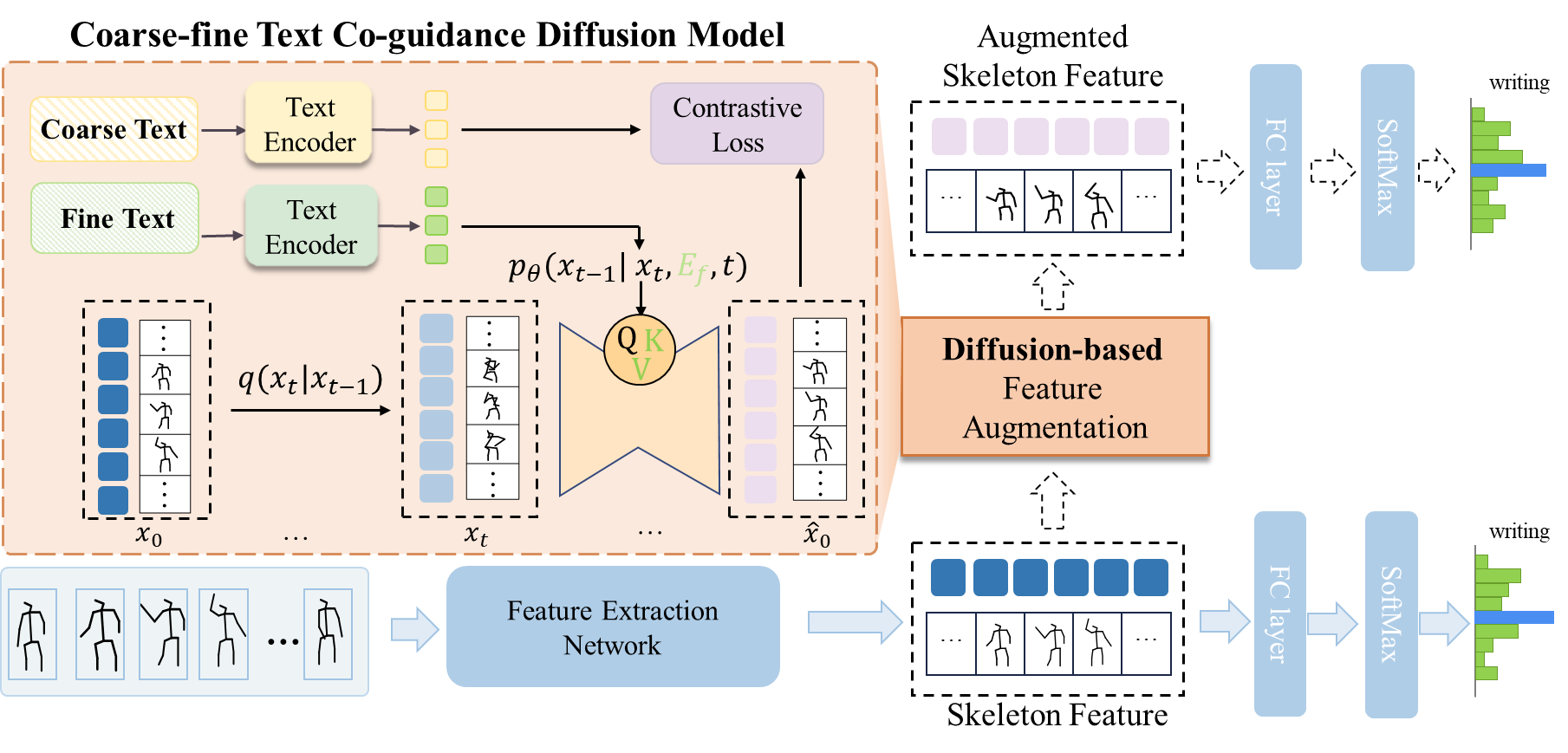}
\caption{Overall framework of the Coarse-fine text Co-guidance Diffusion model (CoCoDiff). During the training phase, high-level skeleton action features are extracted using a feature extraction network and served as inputs to the latent diffusion model for generating diverse action features. Then, the action description as fine text are incorporated as a condition in each iteration of denoising process. And action labels as coarse text embedding is employed to constrain their semantic consistency with generated skeleton features $\bm{{\widehat{x}_0}}$ using contrastive loss. In the inference phase, a well-trained feature extraction network is employed for action classification. }
\label{fig2} 
\end{figure*}

\section{Preliminary}

The diffusion model~\cite{ho2020denoising} defines a diffusion process based on a Markov chain, gradually adding random Gaussian noise to the samples during this process, and then learning to recover clean original samples from the noisy samples in the reverse process. Its forward noise-adding process can be defined as:
\begin{equation}
\begin{aligned}
q(\bm{{x}_t}|\bm{{x}_0} )=\mathcal{N}(\bm{{x}_t};\sqrt{\bar{\alpha}_t}\bm{{x}_0} ,(1-\bar{\alpha}_t)\bm{I}),
\end{aligned}
\label{eq2}
\end{equation}
which transforms data sample $\bm{{x}_0} $  to a latent noisy sample $\bm{{x}_t}$  by adding noise to $\bm{{x}_0} $ for $\bm{t}$ times, $t\in\{0, 1, ...,T\}$. Where $\bar{\alpha}_t=\prod_{s=1}^t(1-\beta_s)$, $\beta_s$ represents the scale of the noise variance. The generation process of the diffusion model can be understood as a reverse denoising process, aiming to approximate the clean data $\bm{{x}_0} $ in the target data distribution by progressively denoising the variable $\bm{{x}_t}$:
\begin{equation}
\begin{aligned}
p_\theta(\bm{x_{t-1}}|\bm{{x}_t})=\mathcal{N}(\bm{x_{t-1}};\mu_\theta(\bm{{x}_t},t), \sigma_t^2\bm{I}).
\end{aligned}
\label{eq3}
\end{equation}
During training, a neural network $f_\theta(\bm{{x}_t}, t)$ is trained to predict $\bm{{x}_0} $ from $\bm{{x}_t}$ by minimizing the following $\ell_2$ loss~\cite{ho2020denoising}:
\begin{equation}
    \mathcal{L}_\text{train} =  \frac{1}{2}|| f_\theta(\bm{{x}_t}, t) - \bm{{x}_0} ||^2.
\end{equation}
At inference stage, data sample $\bm{{x}_0} $ is reconstructed from noise $\bm{{x}_t}$ with the model $f_\theta$ and an updating rule in an iterative way,  $\bm{{x}_T} \rightarrow \bm{{x}_{T-\Delta}} \rightarrow ... \rightarrow \bm{{x}_0} $.
\section{Coarse-fine Text Co-guidance Diffusion Model}

\subsection{Overview}
In this section, we provide a detailed introduction to the proposed method. Our CoCoDiff method, which is a plug-and-play module, utilizes the diffusion model to generate diverse action features, assisting in the training of the feature extraction network to enhance its generalization. As shown in Figure~\ref{fig2}, during the training stage, we employ the GCN~\cite{yan2018spatial,chen2021channel,lee2023hierarchically} to extract skeleton data into corresponding high-level action features, serving as the input for latent diffusion model. Subsequently, we incorporate fine text embedding as a condition into each iteration of denoising process. Following that, we align the coarse text embedding with the generated features $\bm{{\widehat{x}_0}}$ using contrastive loss. This coarse-fine text co-guided strategy not only provides fine-grained guidance for feature generation but also enhances the category attributes of features. It keeps a balance between the diversity and semantic consistency of the generated features. In the inference phase, the diffusion model does not participate and we employ the well-trained feature extraction network to classify action features.

\subsection{Diffusion Model for Action Feature Diversity.}
In this paper, we employ a latent diffusion model~\cite{rombach2022high} to generate diverse skeleton action features. These features not only preserve the semantic information of the original actions but also include a certain amount of randomness. Diffusion model aims to learn a model distribution $\bm{{\widehat{x}_0}}$ to approximate a data distribution $\bm{{x}_0}$ using two mutually inverse processes: the forward process and the reverse process. 

The forward process converts the original action feature (obtained by the feature extraction network) $\bm{{x}_0} $ to a latent noisy sample $\bm{{x}_t}$ for $t\in\{0, 1, ...,T\}$ by adding Gaussian noise to $\bm{{x}_0} $. In our setting, the input $\bm{{x}_0} $ of the diffusion model is a 1D high-level feature vector extracted by the feature extraction network. In the reverse process, a neural network $f_{\theta}(\bm{x_{t}},t,\bm{E_{f}})$ is trained to predict $\bm{{x}_0} $ from noisy sample $\bm{{x}_t}$, conditioned on the corresponding action description embedding $\bm{E_{f}}$. Through the forward and reverse process, the diffusion model generates a new diversity action feature $\bm{{\widehat{x}_0}}$, which is semantically consistent with $\bm{{x}_0} $. However, due to the influence of noise, there are somewhat random differences.
\subsection{Coarse-fine Text Co-guidance}
\begin{figure}[t]
  \centering
   \includegraphics[width=\linewidth]{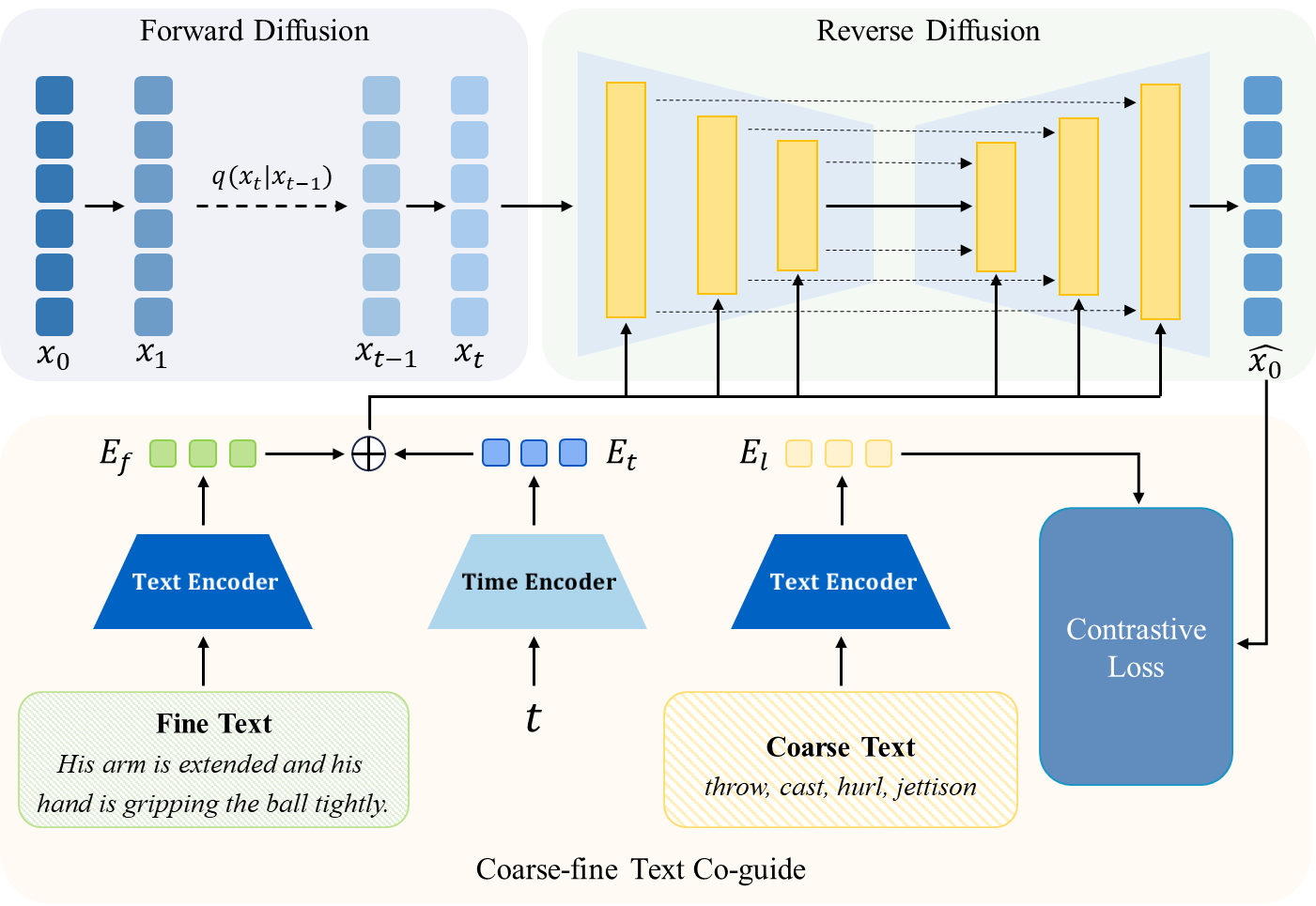}
   \caption{Coarse-fine text co-guided strategy for the action "throw". Two types of textual descriptions for each action are generated using a large language model (GPT-3.5) and using text encoder for feature embedding. Subsequently, the fine text feature is employed to progressively guide the generation of features during the diffusion process. And coarse text feature is utilized for contrastive learning with the generated action features.}
   \label{fig3}
\end{figure}
In order to overcome the limitations of traditional diffusion models in task-orientation and enhance the semantic consistency between generated results and input data, we consider introducing richer guided information. As shown in Figure~\ref{fig3}, we use a large language model to generate two types of textual descriptions for each action: detailed action descriptions and action labels along with their synonyms. Subsequently, we employ these textual descriptions to guide the diffusion model, initially utilizing detailed action descriptions to progressively guide the generation of features during the diffusion process. Following that, we further use action labels for contrastive learning with the generated results, aiding the diffusion model in producing more realistic and semantically enriched action features.

\begin{figure}[t]
  \centering
   \includegraphics[width=\linewidth]{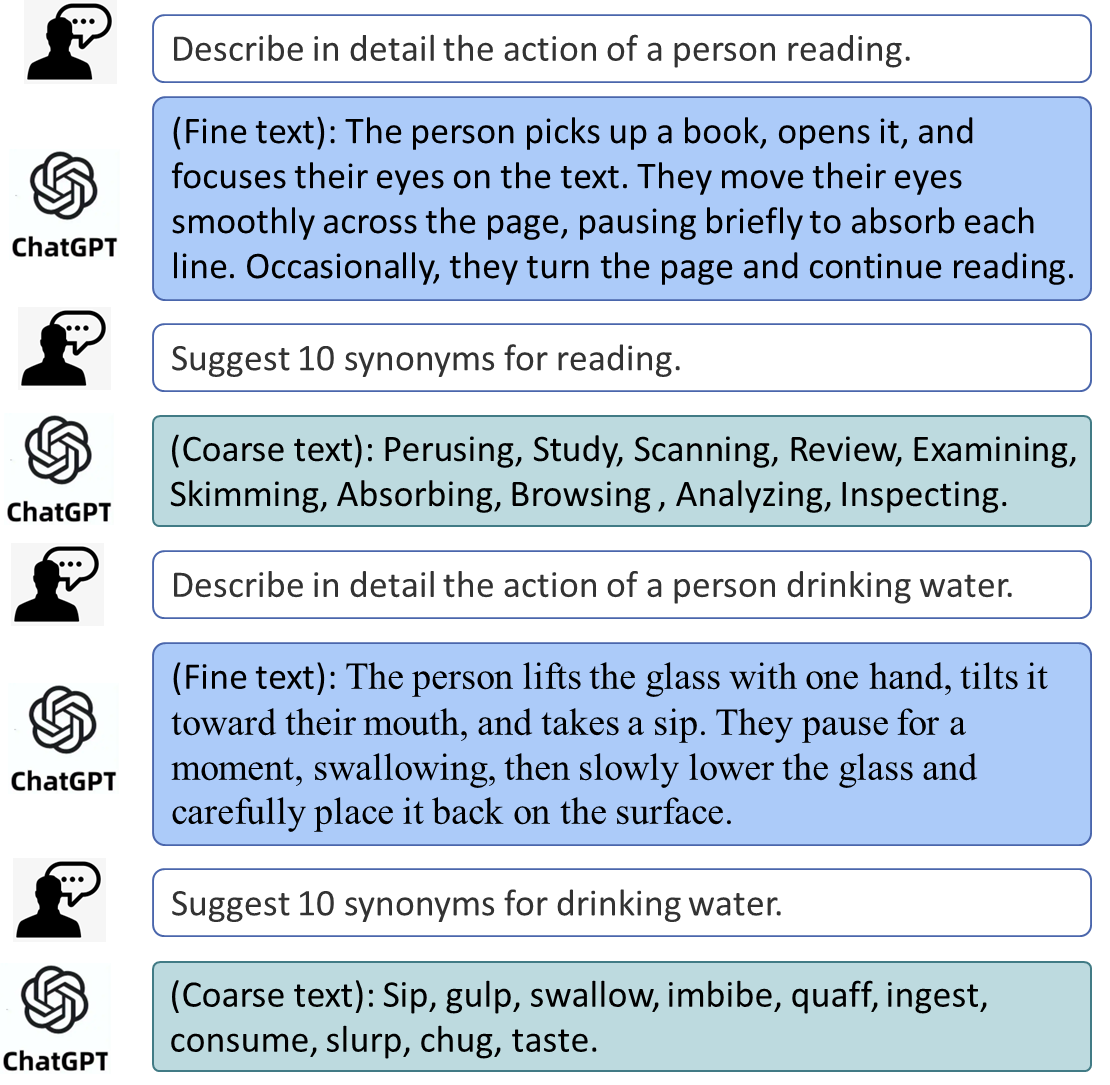}
   \caption{Coarse and fine text generated from different prompt
 inputs by GPT-3.5.}
   \label{fig7}
\end{figure}
\textcolor{black}{\textbf{Text Generation and Encoder.}} We employ GPT-3.5 to generate textual descriptions, which can be categorized into two types: 
i) Fine text: detailed action descriptions. ii) Coarse text: action labels and their synonyms. We take “reading” as an example and present the prompts used for generating coarse and fine text in Figure~\ref{fig7}.

We use a pretrained transformer-based language model~\cite{radford2021learning} as our text encoder to obtain the corresponding text embeddings. The input is in the form of text and undergoes a standard tokenization process. Subsequently, the features are processed through a series of transformer blocks.
The final output is a feature vector that represents the text
description.

\textcolor{black}{\textbf{Fine Text Guidance.}}
For action feature of skeleton sequence, detailed action descriptions can assist the diffusion model in understanding the spatio-temporal information of action features. 
So we incorporate additional text guidance to train a denoising neural network $f_{\theta}(\bm{x_{t}},\bm{E_{t}},\bm{E_{f}})$ of diffusion model and generate $\bm{{\widehat{x}_0}}$ by minimizing the following loss function:
\begin{equation}
\begin{aligned}
\mathcal{L}_{recon}=\frac{1}{2}\left\|f_{\theta}(\bm{x_{t}},\bm{E_{t}},\bm{E_{f}})-\bm{x_{0}}\right\|^{2},
\end{aligned}
\end{equation} 
where $\bm{E_{t}}$ represents the embedding of $t$. $\bm{E_f}$ represents the embedding obtained by encoding the action descriptions information through the text encoder. During the generation phase, the action feature $\bm{{\widehat{x}_0}}$ is reconstructed iteratively from the noisy feature $\bm{x_t}$ using the UNet~\cite{ronneberger2015u} as noise prediction network $f_{\theta}$, namely, $\bm{x_t}$ → $\bm{x_{t-1}}$ →…→ $\bm{{\widehat{x}_0}}$.

\textcolor{black}{\textbf{Coarse Text Guidance.}} After the action descriptions guide diffusion process, the generated skeleton features may still lack semantic consistency in action recognition task. So we further enhance the category attributes of skeleton action features using action label information. We optimize the skeleton-text contrastive loss~\cite{wang2021actionclip} by contrasting skeleton feature representations and action label embeddings in two directions:
\begin{equation}
	\begin{aligned}
    p_{i}^{s2l}(\bm{s_i}) = \frac{\exp(sim(\bm{s_i},\bm{l_{i}})/\tau)}{\sum ^{B} _{j=1} \exp(sim(\bm{s_i},\bm{l_{j}})/\tau)}, \\
    p_{i}^{l2s}(\bm{l_i}) = \frac{\exp(sim(\bm{l_{i}},\bm{s_{i}})/\tau)}{\sum ^{B} _{j=1} \exp(sim(\bm{l_i},\bm{s_{j}})/\tau)},
\end{aligned}
\label{eq:prob}
\end{equation} 
where $\bm{s}$ and $\bm{l}$ represent feature representations of skeleton and label information, respectively. $i$ represents the $i$-th sample in the batch size. $sim(\bm{s}, \bm{l})$ is the cosine similarity, $\tau$ is the temperature parameter, and $B$ is the batch size. Since the number of skeleton sequences is larger than the number of labels, it will inevitably appear that there will be multiple skeleton sequences with the same label in a batch, meaning that $y^{s2l}$ and $y^{l2s}$ may have more than one positive pair. Therefore, instead of using cross-entropy loss, we employ $KL$ divergence as the skeleton-text contrastive loss:
\begin{equation}
	\begin{aligned}
    \mathcal{L}_{con} = \frac{1}{2} E_{\bm{s},\bm{l} \sim \mathcal{D}}  [KL(p^{s2l}(\bm{s}), y^{s2l}) + KL(p^{l2s}(\bm{l}), y^{l2s})],
\end{aligned}
\label{eq:kl}
\end{equation}
where $D$ represents the entire dataset, $y^{s2l}$ and $y^{l2s}$ are the label values of similarity scores. When the skeleton embedding $\bm{s}$ matches the label embedding $\bm{l}$, the value of $y^{s2l}$ and $y^{l2s}$ are $1$.

\begin{algorithm}[t]
   \caption{$\textbf{Training Algorithm of CoCoDiff}$}
\label{alg:train}
\begin{algorithmic}[1]
    \STATE \textbf{Stage 1:}  pretraining of diffusion model
    \STATE {\bfseries Input:}
    feature X $\in$ $R^{B \times C}$ extrated by a trained GCN , coarse text embedding $E_l$ $\in$ $R^{B \times N}$, fine text embedding $E_f$ $\in$ $R^{B \times N}$
    \FOR{all $\bm{x_0} \in X$}
        \STATE Sample $t \sim [1, T]$; \  $\epsilon \sim \mathcal{N}(0, I)$
        \STATE $\mathbf{x}_t = \sqrt{\bar \alpha_{t}}\mathbf{x} + \sqrt{1-\bar\alpha_{t}} \epsilon$
        \STATE Compute $\mathcal{L}_{recon}$ loss by Eq.(4) and compute $\mathcal{L}_{con}$ loss by Eq.(6)
        \STATE Update with  $\mathcal{L}_{diff}$ = $\mathcal{L}_{recon}$ + $\lambda$$\mathcal{L}_{con}$ 
    \ENDFOR
    \STATE \textbf{Stage 2:}  training of GCN network
    \STATE {\bfseries Input:}
    Skeleton sequence S $\in$ $R^{B \times C \times T \times V \times M}$, coarse text embedding $E_l$ $\in$ $R^{B \times N}$, fine text embedding $E_f$ $\in$ $R^{B \times N}$
    \STATE Feature X $\in$ $R^{B \times D}$ extrated by GCN
    \FOR{all $\bm{x_0} \in X$}
        \STATE Sample $t \sim [1, T]$; \  $\epsilon \sim \mathcal{N}(0, I)$
        \STATE $\mathbf{x}_t = \sqrt{\bar \alpha_{t}}\mathbf{x} + \sqrt{1-\bar\alpha_{t}} \epsilon$
        \STATE Compute $\mathcal{L}_{diff}$ loss by Eq.(7) and compute $\mathcal{L}_{cls}$ loss by Eq.(8)
        \STATE Update with  $\mathcal{L}$ = $\mathcal{L}_{cls}$ + $\mathcal{L}_{diff}$
    \ENDFOR
\end{algorithmic}
\end{algorithm}

\subsection{Training and Inference}
\textcolor{black}{\textbf{Training.}} In order to improve training efficiency and help the model capture the semantic information of skeleton data, we divide the training strategy into two stages. The training process can be summarized in the following steps:

1) Using a pre-trained GCN to generate action features, which are then used to train the diffusion model. In the first training stage, we conduct the training of the diffusion model. Firstly, we initially utilize the original GCN to generate high-level action feature representations, serving as the input for the diffusion model. Subsequently, the training process of the diffusion model is guided through action description and label information, aiming for a more profound learning of action features.

2) Using the trained diffusion model to generate enhanced features, which are then used to train a new randomly initialized GCN action recognition model. The second training stage involves the training of the GCN and the fine tune for diffusion model, utilizing the feature representation $\bm{x_0}$ extracted by the GCN as input. In this phase, we employ the pre-trained diffusion model to generate action features of the same category, denoted as $\bm{{\widehat{x}_0}}$. Then, $\bm{x_0}$ and $\bm{\widehat{x_0}}$ are fed into the classification network. Algorithm~\ref{alg:train} provides the pseudo-code of CoCoDiff training procedure.

\textcolor{black}{\textbf{Loss Functions.}}
For the first stage of training, the loss function is crucial in generating diverse and effective action feature. This loss function, denoted as $\mathcal{L}_{diff}$, is composed of two parts: $\mathcal{L}_{recon}(\bm{E_{s}},\bm{E_{f}})$ and $\mathcal{L}_{con}(\bm{E_{s}},\bm{E_{l}})$. $\mathcal{L}_{recon}(\bm{E_{s}},\bm{E_{f}})$ represents the reconstruction loss, measuring the difference between the skeleton action feature generated under detailed action description and the original skeleton action feature. It guides the model to reconstruct the original features as accurately as possible. The skeleton-text contrastive loss ($\mathcal{L}_{con}(\bm{E_{s}},\bm{E_{l}})$) controls the semantic consistency between skeleton features and corresponding action label information by computing the similarity between skeleton features and action label embeddings. It assists the model in aligning the generated skeleton features with the associated action labels. Therefore, the loss in the first stage can be represented as:
\begin{equation}
\begin{aligned}
\mathcal{L}_{diff}=\mathcal{L}_{recon}(\bm{E_{s}},\bm{E_{f}})+\lambda \mathcal{L}_{con}(\bm{E_{s}},\bm{E_{l}}),
\end{aligned}
\label{eq:diff}
\end{equation} 
where $E_s$ represents the skeleton action feature representation obtained by the GCN, $E_l$ is the embedding of action label information encoded by the text encoder, and $\lambda$ is a trade-off parameter.

In the training of the second stage, the cross-entropy loss $\mathcal{L}_{cls}$
is employed to assess the difference between the classification results and the actual one-hot ground-truth action labels:
\begin{equation}
\begin{aligned}
\mathcal{L}_{cls}=-y\log p_{\theta}(\bm{x}),
\end{aligned}
\end{equation} 
where $y$ is the one-hot ground-truth action label, $p_{\theta}(\bm{x})$ is the predicted probability distribution.

Finally, $\mathcal{L}_{cls}$ loss is combined with the first training stage loss $\mathcal{L}_{diff}$ to form the full learning objective function:
\begin{equation}
\begin{aligned}
\mathcal{L}=\mathcal{L}_{cls}+ \mathcal{L}_{diff}.
\end{aligned}
\end{equation} 
\textcolor{black}{\textbf{Inference.}} As shown in Figure~\ref{fig2}, 
although we divide training process into two stages,  and utilize two loss functions jointly to train the model. The diffusion model does not participate in the inference process, the input of classification network consists solely of $\bm{x_0}$. After training is completed, we use the skeleton sequence as input, extract action features using the well-trained GCN, and then classify them.
\section{Experiments}
\subsection{Datasets}
\textcolor{black}{\textbf{NTU RGB+D}}~\cite{shahroudy2016ntu} is a widely used dataset, containing 56,880 samples. Forty participants were invited to perform 60 types of actions, including daily activities and health-related movements. Each action was completed by 1 to 2 persons. The human skeleton is represented by 25 3D joints captured by Microsoft Kinect v2 cameras set at three different horizontal angles. The dataset’s standard evaluation protocols, including cross-subject (X-Sub) and cross-view (X-View). The X-Sub protocol is divided based on subjects. The training set consists of 20 subjects, and the test set consists of other 20 subjects. The X-View protocol is segmented based on camera views. Camera views 2 and 3 are chosen to construct the training data, while camera view 1 is used for testing.

\textcolor{black}{\textbf{NTU RGB+D 120}}~\cite{liu2019ntu} extends NTU RGB+D by introducing 60 new action classes, adding 57,367 samples. It collects a total of 120 action classes completed by 106 participants, comprising 113,945 samples. It also increases the number of camera setups to 32 by using different locations and backgrounds. Two evaluation protocols, namely cross-subject (X-Sub) and cross-setup (X-Set), are recommended. In the X-Sub, samples are selected from 56 subjects to form the training set and another 50 subjects for testing. In the X-Set, samples with even IDs are used for training, while those with odd IDs are used for testing.

\textcolor{black}{\textbf{Kinetics-Skeleton}}~\cite{kay2017kinetics} is come from the Kinetics 400 video dataset, utilizing the
OpenPose pose estimation~\cite{cao2017realtime} to extract 240,436 training data and 19,796 testing skeleton sequences across 400 classes. The dataset limits each time step to only two skeletons and filters out skeletons with lower confidence scores. This process guarantees high-quality sequences for research on human action recognition and pose estimation.

\subsection{Implementation Details}
We adopt GCN~\cite{chen2021channel,lee2023hierarchically,zhou2024blockgcn} as the backbone and implement the proposed method using the PyTorch deep learning framework. All experiments are conducted on an RTX 4090 GPU. We use the stochastic gradient descent optimizer with a momentum of 0.9 and weight decay of 0.0004 to train the model. A warm-up strategy is employed for stable training in the first 5 epochs. In the ablation studies, the initial learning rate is set to 0.1, and we decay it by a factor of 0.1 from epoch 70 to 80. We train the models for 90 epochs and select the best-performing one. For NTU RGB+D~\cite{shahroudy2016ntu} and NTU RGB+D 120~\cite{liu2019ntu}, we set the batch size to 64, and all samples are adjusted to 64 frames. For Kinetics-Skeleton, the batch size is set to 128. In addition,
to overcome the absence of neck node in the Kinetics-Skeleton~\cite{kay2017kinetics}, we define the center of both shoulder joints as neck node, resulting in a total of 18 nodes.

\subsection{Ablation Study}
\begin{table}[t]
 \caption{Performance of the proposed method using different GCN-based backbones on NTU RGB+D and NTU RGB+D 120 dataset with the joint input modality.}
  
  \begin{center}
  \begin{small}
  \begin{tabular}{l c c c}
  \toprule
  \multirow{2}{*}{ \textbf{Method} }  & \multicolumn{2}{c} {\textbf{NTU RGB+D 120} } \\
     		& \textbf{X-Sub (\%)} & \textbf{X-Set (\%)} \\
  \midrule
  \midrule
  CTR-GCN~\cite{chen2021channel}  & 84.9 & 85.9 \\
  ~~+ CoCoDiff  & $85.8^{\uparrow 0.9}$ & $86.5^{\uparrow 0.6}$ \\
  \midrule
  HD-GCN~\cite{lee2023hierarchically}   & 85.1 & 86.7 \\
  ~~+ CoCoDiff  & $86.0^{\uparrow 0.9}$ & $87.4^{\uparrow 0.7}$ \\
   \midrule
  Block-GCN~\cite{zhou2024blockgcn}   & 86.3 & 88.1 \\
  ~~+ CoCoDiff  & $86.9^{\uparrow 0.6}$ & $88.7^{\uparrow 0.6}$ \\
  \bottomrule
\end{tabular}
\end{small}
\end{center}
\centering
 
  \label{tab::Backbones}
\end{table}

\textbf{Effectiveness of CoCoDiff on Different Backbones.} The module we propose is plug-and-play, compatible with most GCN-based backbone networks. To examine its effectiveness, we applly it to three widely-used GCN-based backbone networks~\cite{chen2021channel,lee2023hierarchically,zhou2024blockgcn} and evaluate it on the NTU RGB+D 120 dataset. For a fair comparison, the training parameters (learning rate, weight decay, etc.) of each GCN are kept entirely consistent with the settings after the addition of CoCoDiff during training. Table \ref{tab::Backbones} presents the performance of CoCoDiff on different backbone, and we can observe that all models obtained significant accuracy gains by using CoCoDiff. 
\begin{table}[t]
 \caption{Text guided method exploration of the proposed method on
NTU-RGB+D 120 dataset under the X-Sub setting with the joint
input modality. The best one is in \textbf{bold}.}
  
  \begin{tabular}{c c c c c}
  \toprule
  \multirow{2}{*}{\textbf{CTR-GCN}} & \multirow{2}{*}{\textbf{Diffusion}} & \multicolumn{2}{c}{\textbf{Text Guidance}} & \multirow{2}{*}{\textbf{Acc (\%)}} \\
  \cmidrule{3-4}
                &   &   \textbf{Fine }   &     \textbf{Coarse}     &       \\ 
  \midrule
  \midrule
  $\checkmark$ &$\times$ &$\times$ &$\times$ & 84.9\\
  $\checkmark$ &$\checkmark$ &$\times$ &$\times$ & 85.1\\ 
  $\checkmark$ &$\checkmark$ &$\checkmark$ &$\times$ & 85.5 \\
  $\checkmark$ &$\checkmark$ &$\times$ &$\checkmark$ & 85.4 \\
  $\checkmark$ &$\checkmark$ &$\checkmark$ &$\checkmark$ & \textbf{85.8 }\\
  \bottomrule
\end{tabular}
\centering
 
 \label{tab::Text Guide}
\end{table}

\textbf{Influences of Text Guidance.} We show the influences of different text guided methods in Table \ref{tab::Text Guide}. By directly using diffusion model without text guide, the model only slightly outperforms (0.2\%) the baseline CTR-GCN~\cite{chen2021channel}. Utilizing the fine text (action description) or coarse text (action label and synonyms) could improve the performance (0.6\% and 0.5\%). Using the coarse-fine text co-guidance leads to the performance with 0.9\% improvement. 

The experimental results indicate that text guidance can provide the model with more abundant semantic information. We speculate the reason is that fine text provides more precise action descriptions and richer semantic details, facilitating the model in generating actions more accurately. Besides, coarse text may offer broader semantic context, enabling the model to comprehend action content more comprehensively. Moreover, the combination of coarse and fine text can be used to provide richer and more accurate semantic information, thus improving model performance.

\textbf{Influences of $\lambda$ Selection.} To study the influences of trade-off parameter $\lambda$ in Eq.~\ref{eq:diff}, we search the value of $\lambda$  with 5-fold cross-validation on CTR-GCN~\cite{chen2021channel}. The $\lambda$ is in $\{0.8,0.075, 0.01, 0.002, 0.001\}$. The performance of models are $84.7\%$, $85.8\%$, $85.4\%$, $85.1\%$ and $84.8\%$, respectively.  When $\lambda = 0.075$, namely CoCoDiff using text information achieves the best performance on CTR-GCN ($85.8\%$). Compared to the baseline, it improves $0.9\%$. It seems that bigger $\lambda$ may hurt the performance while too small values only provide a little improvement. Finally, we choose the configuration of $\lambda = 0.075$ for the following experiments.

\begin{table}[t]
  \caption{Hyper-parameter exploration of the proposed method on
NTU-RGB+D 120 dataset under the X-Sub setting with the joint
input modality. The best one is in \textbf{bold}.}
  
  \begin{tabular}{l c c}
  \toprule
  \textbf{Method} & $\lambda$ & \textbf{Acc (\%)} \\
  \midrule
  \midrule
  CTR-GCN~\cite{chen2021channel} & - & 84.9 \\
  \midrule
  CTR-GCN+ CoCoDiff & $8\times10^{-1}$ & 84.7 \\
  CTR-GCN+ CoCoDiff & $7.5\times10^{-2}$ & \textbf{85.8} \\
  CTR-GCN+ CoCoDiff & $1\times10^{-2}$ & 85.4 \\
  CTR-GCN+ CoCoDiff & $1\times10^{-3}$ & 85.1\\
  CTR-GCN+ CoCoDiff & $2\times10^{-3}$ & 84.8\\
    
  \bottomrule
\end{tabular}
\centering

 \label{tab::Hyper-parameter}
\end{table}

\begin{table}[t]
\centering
  \caption{Diffusion steps exploration of the proposed method on
NTU-RGB+D 120 dataset under the X-Sub setting with the joint
input modality. The best one is in \textbf{bold}.}
  \label{tab::Diffusion steps}
  \begin{tabular}{l c c}
  \toprule
  \textbf{Method} & \textbf{T} & \textbf{Acc (\%)} \\
  \midrule
  \midrule
  CTR-GCN~\cite{chen2021channel} & - & 84.9 \\
  \midrule
  CTR-GCN+ CoCoDiff & 10 & 84.7 \\
  CTR-GCN+ CoCoDiff & 20 & 84.8 \\
  CTR-GCN+ CoCoDiff & 25 & 84.9 \\
  CTR-GCN+ CoCoDiff & \textbf{30} & \textbf{85.8} \\
  CTR-GCN+ CoCoDiff & 35 & 84.9\\
  CTR-GCN+ CoCoDiff & 40 & 84.7\\
  \midrule 
  HD-GCN~\cite{lee2023hierarchically} & - & 85.1 \\
  \midrule
  HD-GCN+ CoCoDiff & 5 & 85.2\\
  HD-GCN+ CoCoDiff & \textbf{10} & \textbf{86.0}\\
  HD-GCN+ CoCoDiff & 20 & 85.4\\
  HD-GCN+ CoCoDiff & 25 & 85.3\\
    
  \bottomrule
\end{tabular}
\end{table}

\begin{table}[t]
\centering
  \caption{Training strategy exploration of the proposed method
on NTU-RGB+D 120 dataset under the X-Sub setting with the
joint input modality. The best one is in \textbf{bold}.}
  \label{training strategy}
  \vskip 0.15in
  \begin{tabular}{c c c }
  \toprule
  \textbf{Pretrain GCN} &\textbf{Pretrain Diffusion} & \textbf{Acc (\%)} \\
  \midrule
  \midrule
  $\checkmark$ & $\checkmark$ & \textbf{85.8} \\
  $\times$ & $\times$ & 85.1 \\
  $\times$ & $\checkmark$ & 85.4\\
 
  \bottomrule
\end{tabular}
\vskip 0.15in
\end{table}

\begin{table}[t]
\centering
  \caption{Comparison of training time of the proposed method
on NTU-RGB+D 120 dataset under the X-Sub setting with the
joint input modality. }
  \label{tab::model size}
  \vskip 0.15in
  \begin{tabular}{l c c}
  \toprule
  \textbf{Method}  & \textbf{Training time(min)} & \textbf{Acc(\%)} \\
  \midrule
  \midrule
  CTR-GCN & 9min 2s & 84.9 \\
  \midrule
    CTR-GCN+GAP~\cite{xiang2023generative} & 11min 25s & 85.5 \\
    CTR-GCN+FR Head~\cite{zhou2023learning} &10min 11s & 85.5 \\
    CTR-GCN+ E-Koopman~\cite{wang2023neural} & 12min 25s & 85.7 \\
    CTR-GCN+CoCoDiff & 12min 20s & \textbf{85.8} \\
    
  \bottomrule
\end{tabular}
\end{table}
\begin{table*}[t]
  \caption{Performance comparison of skeleton-based action recognition in top-1 accuracy (\%). The best one is in \textbf{bold} and the second one is \underline{underlined}.}
  
  \resizebox{\textwidth}{!}{%
   \begin{tabular}{l c c c c c c c}
    \toprule
     \multirow{2}{*}{ \textbf{Method} } & \multirow{2}{*}{ \textbf{Publication} } & \multicolumn{2}{c}{ \textbf{NTU RGB+D} } & \multicolumn{2}{c}{ \textbf{NTU RGB+D 120} } & \multicolumn{2}{c}{ \textbf{Kinetics-Skeleton} }  \\
     		& & \textbf{X-Sub(\%)} & \textbf{X-View(\%)} & \textbf{X-Sub(\%)} & \textbf{X-Set(\%)} & \textbf{Top-1(\%)} & \textbf{Top-5(\%)}\\
    \midrule
    ST-GCN~\cite{yan2018spatial} & AAAI2018 & 86.1 & 90.1 & 82.1 &84.5 &30.7 &52.8 \\
    2s-AGCN~\cite{shi2019two} & CVPR2019 & 88.5 & 95.1 & 82.9& 84.9 &36.1 &58.7 \\
    SGN~\cite{SGN2020} & CVPR2020 & 89.0 & 94.5 & 79.2 &81.5 & - & - \\
    Shift-GCN~\cite{Shift-GCN2020} & CVPR2020 & - & - & 80.9 & 83.2 & - & - \\
    MS-G3D~\cite{MS-G3D-2020} & CVPR2020 & 89.6 & 95.0 & 84.0 &86.0 &38.0 &60.9 \\
    DC-GCN+ADG~\cite{DC-GCN+ADG2020} & ECCV2020 & - & - & 82.4 & 84.3 & - & -\\
    MST-GCN~\cite{MST-GCN-2021} & AAAI2021 & 89.0 & 95.1 & 82.8 &84.5 &38.1 &60.8 \\
    CTR-GCN~\cite{chen2021channel} & ICCV2021 & 89.6 & 95.2 & 84.9 &85.9 & - & -  \\
    STF~\cite{STF2022} & AAAI2022 & - & - & 85.0 &86.4 &\underline{39.9} & - \\
   Info-GCN~\cite{InfoGCN-2022} & CVPR2022 & 89.8 & 95.2 &  85.1 &86.3 & - & -\\
    HD-GCN~\cite{lee2023hierarchically} & ICCV2023 & {90.0} & \underline{95.4} &  85.1 & 86.7 & 38.9 & \underline{61.7}\\
    SPIANet~\cite{yin2024spatiotemporal}&PR2024  & 90.3 &{95.2}& {85.4} & {87.0} & - & -\\
    BlockGCN~\cite{zhou2024blockgcn}  &CVPR2024& \underline{90.8} &{95.2}
    & {86.3} & \underline{88.1} & - & -\\
    LA-GCN~\cite{xu2023language}  &TMM2025& -  & - 
    & \underline{86.5} & {88.0} & - & -\\
    \midrule
    \textbf{Ours}   &-& \textbf{90.9} &\textbf{95.6}  & \textbf{86.9} &\textbf{88.7} & \textbf{41.4}&\textbf{65.4} \\
    \bottomrule
\end{tabular}
}
\centering

  \label{tab::ALL}
\end{table*}
\textbf{Influences of $T$ Selection.}  In the diffusion model, we refer to the parameter T as "diffusion steps", denoting the total number of steps for progressively adding noise to samples during the forward diffusion process. This parameter serves a pivotal role in controlling the level of noise injected into the samples, thereby impacting the diversity and quality of the generated samples. To investigate the impact of this crucial parameter on model performance, we attempt to explore the influence of different diffusion steps across various backbones, and the results are available in Table~\ref{tab::Diffusion steps}. On CTR-GCN~\cite{chen2021channel}, we tested different values of T to assess their impact on model performance. We found that when the value of T is between 25 and 35, the action features generated by CoCoDiff positively contribute to the model. However, once beyond this range, the generated features have a negative impact on the model's performance. After replacing the backbone with HD-GCN~\cite{lee2023hierarchically}, this effective range changes to 10 to 20. 

The occurrence of this phenomenon may involve the following two reasons. First, the differences in model architecture among different backbones lead to variations in the extracted action features, and these features exhibit varying sensitivities to noise. Consequently, the optimal number of noise steps required varies for different backbones to achieve the best performance. Second, under an appropriate number of noise steps, the action features generated by CoCoDiff can better assist the model in learning useful information and thus enhancing the model's generalization performance. While, when the number of noise steps is smaller, the generated features may contain excessive noise, disrupting the semantic information in the original action features and adversely affecting the performance of model. On the other hand, when the number of noise steps is larger, the generated features become too similar to the original features, resulting in a loss of diversity and an inability to improve the generalization performance of model.

\textbf{Influences of training strategy.} We conduct ablation studies on CTR-GCN~\cite{chen2021channel} to validate the effectiveness of the two-stage training strategy proposed in Section \textbf{4.4} of the paper, and the results are available in Table \ref{training strategy}. The training strategies are categorized into three types: the first type utilizes a pretrained GCN network to provide action features for the diffusion model (85.8\%), the second type uses an untrained GCN network for the same purpose (85.4\%), and the third type directly proceeds with the second-stage training (85.1\%). The experimental results demonstrate that the first type of training strategy achieved the best results. We speculate that the action features provided by the pretrained GCN network eliminate much of the irrelevant noise, thereby providing high-quality training data for the pretraining of the diffusion model.

 \textbf{Comparison of Training Time.} In Table~\ref{tab::model size}, we compare the training time of one epoch among several plug-and-play auxiliary training methods similar to CoCoDiff. Since both CoCoDiff and E-Koopman~\cite{wang2023neural} employ a two-stage training approach, we choose to compare training time rather than model size. To ensure a fair comparison, we use the same backbone when comparing training times and keep relevant parameters that may affect training time consistent, such as batch size. For CoCoDiff and E-Koopman~\cite{wang2023neural}, we sum the training times of two stages to obtain the final result. According to the experimental results, the training time of CoCoDiff is slightly higher than that of GAP\cite{xiang2023generative} and FR-Head\cite{zhou2023learning}, comparable to E-Koopman~\cite{wang2023neural}. However, CoCoDiff achieved the highest accuracy on CTR-GCN. This indicates that CoCoDiff achieves a good balance between accuracy and training time.

\begin{figure}[h]
  \centering
   \includegraphics[width=\linewidth]{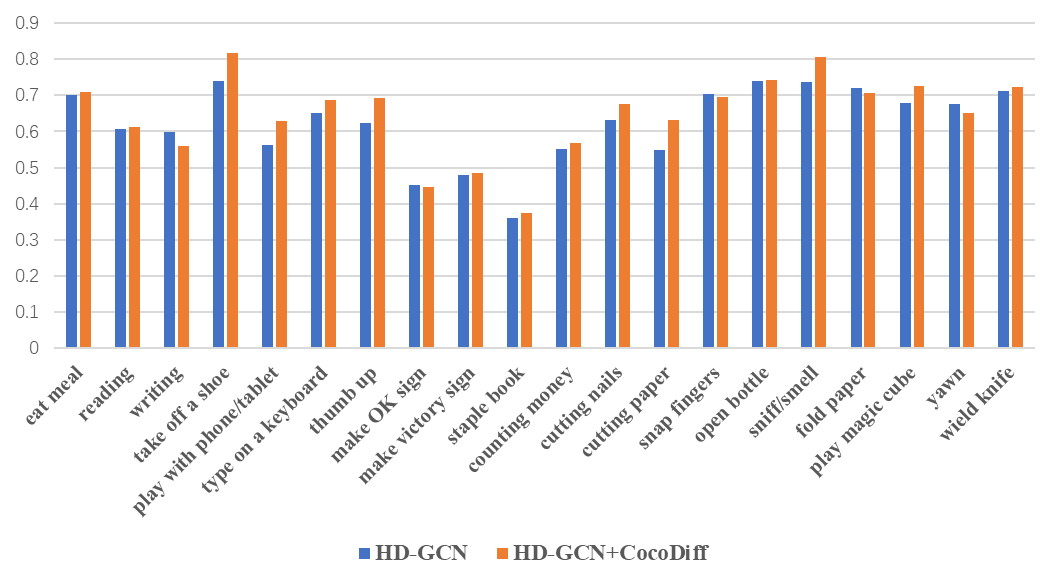}
   \caption{The group-wise accuracy difference ($\%$) between our method and HD-GCN on large intra-class variations actions for NTU RGB+D 120 dataset under the X-Sub setting with the joint input modality.}
   \label{fig4}
\end{figure}

\subsection{Comparison}

In this section, we conduct a comparison with the state-of-the-art methods on NTU RGB+D, NTU RGB+D 120 and  Kinetics-Skeleton datasets to demonstrate the competitive ability of our proposed method.  The quantitative results are displayed in Table~\ref{tab::ALL}. It can be seen that our method outperforms all of the existing methods on these three datasets. On NTU-RGB+D, NTU-RGB+D 120 and Kinetics-Skeleton, our model achieves the best accuracy.

We can observe that compared to NTU-RGB+D, CoCoDiff has achieved higher accuracy improvements on NTU-RGB+D 120 and Kinetics-Skeleton. We speculate that one of the reasons for this result is that with the increase of the data, the corresponding number of samples we generate will also increase, so our method works better on larger datasets. Notably, our method utilizes diffusion model to generate action features for enhancing model generalization performance. The proposed method is not limited to different types of backbones, demonstrating extensive application potential in the field of action recognition.

\subsection{Visualization}
In order to visually demonstrate the improvement of the model's generalization performance by CoCoDiff. Specifically, we use the results of HD-GCN on the NTU RGB+D 120 dataset as the baseline, considering categories with recognition accuracy below 75$\%$ as those with large intra-class gaps. Then, we compare the accuracy of the original HD-GCN~\cite{lee2023hierarchically} on these categories with that of HD-GCN+CoCoDiff. Experiments are conducted on the X-sub subset, with only 3D joint coordinates used as input, and the experimental results are shown in Figure~\ref{fig4}. From the Figure~\ref{fig4}, it can be seen that our proposed CoCoDiff method achieves better results. 

\begin{figure}[ttt]

  \centering
   \includegraphics[width=0.85\linewidth]{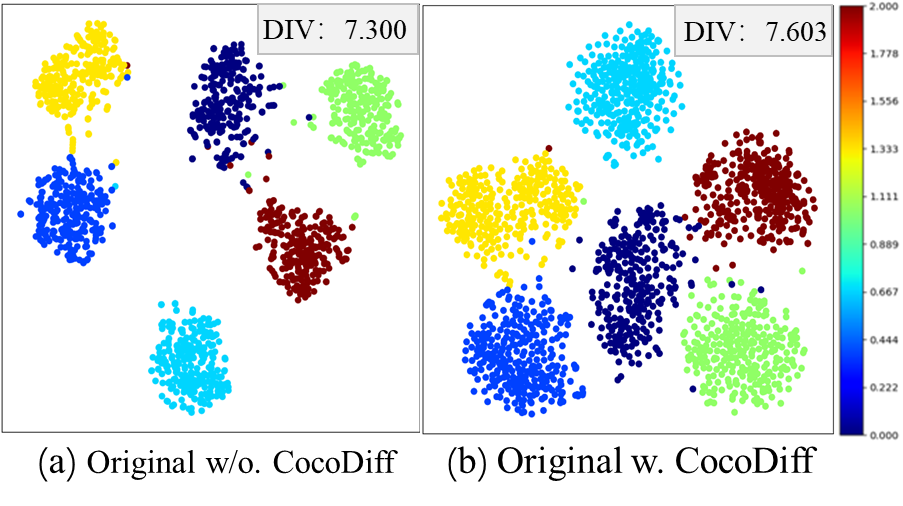}

  \caption{Visualization of latent representation by t-SNE for randomly selected action categories with large intra-class variations from NTU RGB+D 120 dataset. Different colors indicate different classes. (a) is from the original HD-GCN, (b) is the action features generated by CoCoDiff applied on HD-GCN.}
   \label{fig5}
   
\end{figure}

To visually demonstrate the effectiveness of CoCoDiff, we randomly selecte some action categories and visualized their distribution in the feature space using t-SNE~\cite{van2008visualizing}.  Meanwhile, we use Diversity (DIV)~\cite{guo2020action2motion} to calculate variance through action features. It is worth noting that to improve the confidence of the results, the calculated DIV results are derived from all the extracted features. For example, Figure~\ref{fig5} (a) shows the action features extracted by the original HD-GCN on the categories with large intra-class gaps, but the DIV value is calculated from the action features of all 120 categories extracted by the original HD-GCN. Through comparing the DIV values in Figure~\ref{fig5} (a)(7.300) and (b)(7.603), the result demonstrates that CoCoDiff provides HD-GCN with diverse training features.

\section{Conclusion}
In this paper, we propose a Coarse-fine text co-guidance Diffusion model (CoCoDiff) to enhance the generalization of the model by increasing feature diversity in the latent space. Specifically, CoCoDiff employs the latent diffusion model, aiding in model training, to generate high quality and diverse action features. Besides, to ensure semantic consistency between input data and generated results, coarse-fine text co-guided strategy is designed. The proposed model could be applied in a plug-and-play manner to existing action recognition models without extra inference costs. Extensive experiments demonstrate that our proposed method achieves state-of-the-arts performance on skeleton-based action recognition benchmarks, including NTU RGB+D, NTU RGB+D 120 and  Kinetics-Skeleton.

\bibliography{references.bib}

\bibliographystyle{IEEEtran}
\vskip -2\baselineskip plus -1fil

\begin{IEEEbiography}
[{\includegraphics[width=1in,height=1.25in,clip,keepaspectratio]{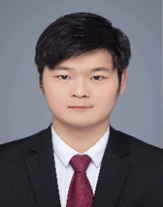}}] 
{Zhifu Zhao} is a Associate Professor in the School of Artificial Intelligence at Xidian University, Xi’an, China. He received his Ph.D. degree in School of Artificial Intelligence from Xidian University in 2020. His research interests are deep learning, video understanding and compressive sensing.
\end{IEEEbiography}
\vskip -2\baselineskip plus -1fil

\begin{IEEEbiography}
[{\includegraphics[width=1in,height=1.25in,clip,keepaspectratio]{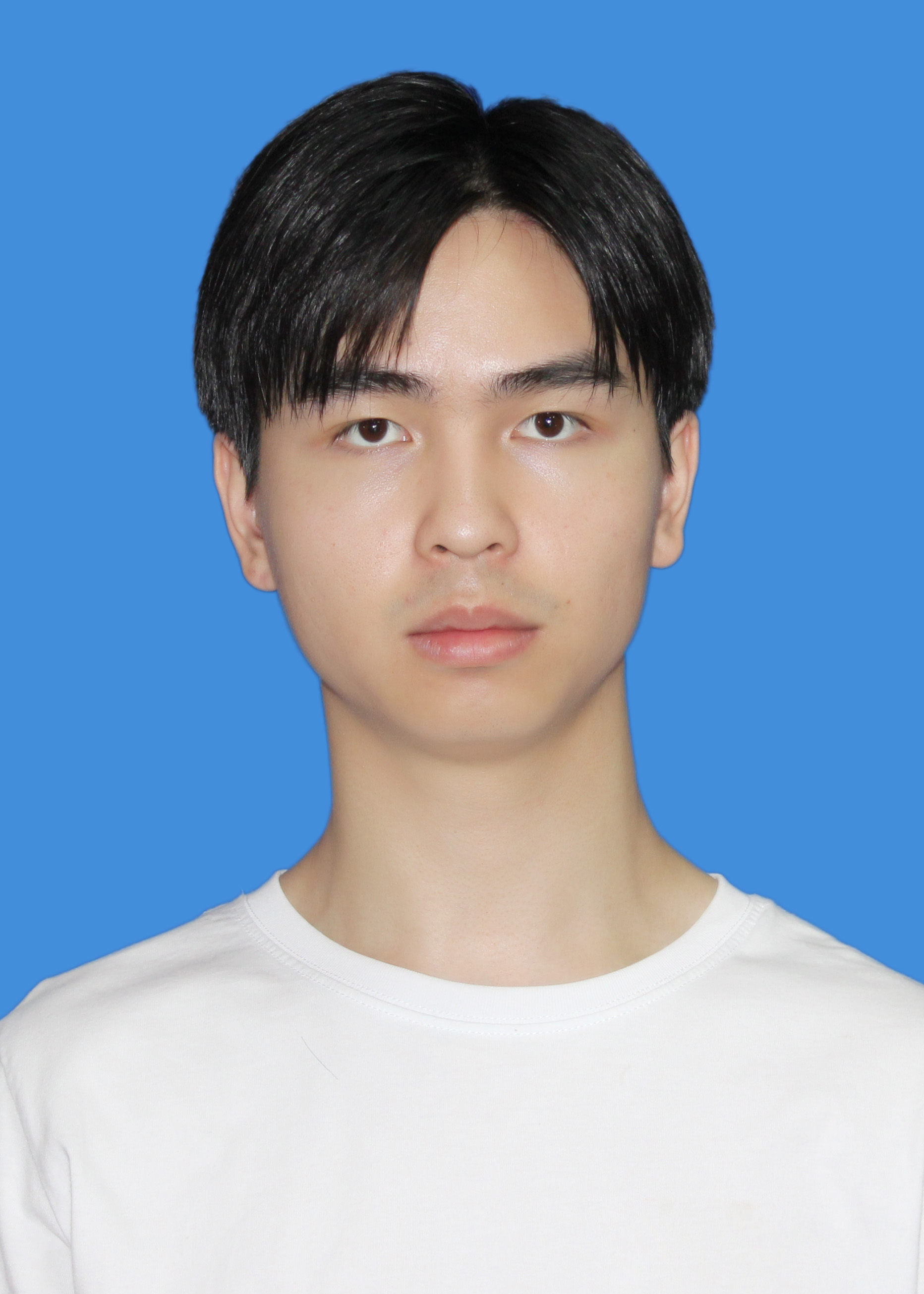}}] 
{Hanyang Hua} received his Bachelor of Engineering (B.E.) degree in Communication Engineering from Chang'an University, Xi'an, China in 2022. He is currently pursuing the master's degree with the Guangzhou Institute of Technology at Xidian University, China. His current research interests are computer vision and action recognition.
\end{IEEEbiography}
\vskip -2\baselineskip plus -1fil

\begin{IEEEbiography}
[{\includegraphics[width=1in,height=1.25in,clip,keepaspectratio]{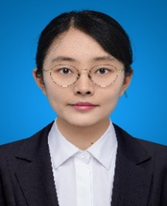}}] 
{Jianan Li} is a Lecturer in the School of Computer Science and Technology at Xidian University, Xi’an, China. She received her Ph.D. degree in School of Artificial Intelligence from Xidian University in 2020. Her research interests are deep learning, video understanding and action recognition.
\end{IEEEbiography}
\vskip -2\baselineskip plus -1fil

\begin{IEEEbiography}
[{\includegraphics[width=1in,height=1.25in,clip,keepaspectratio]{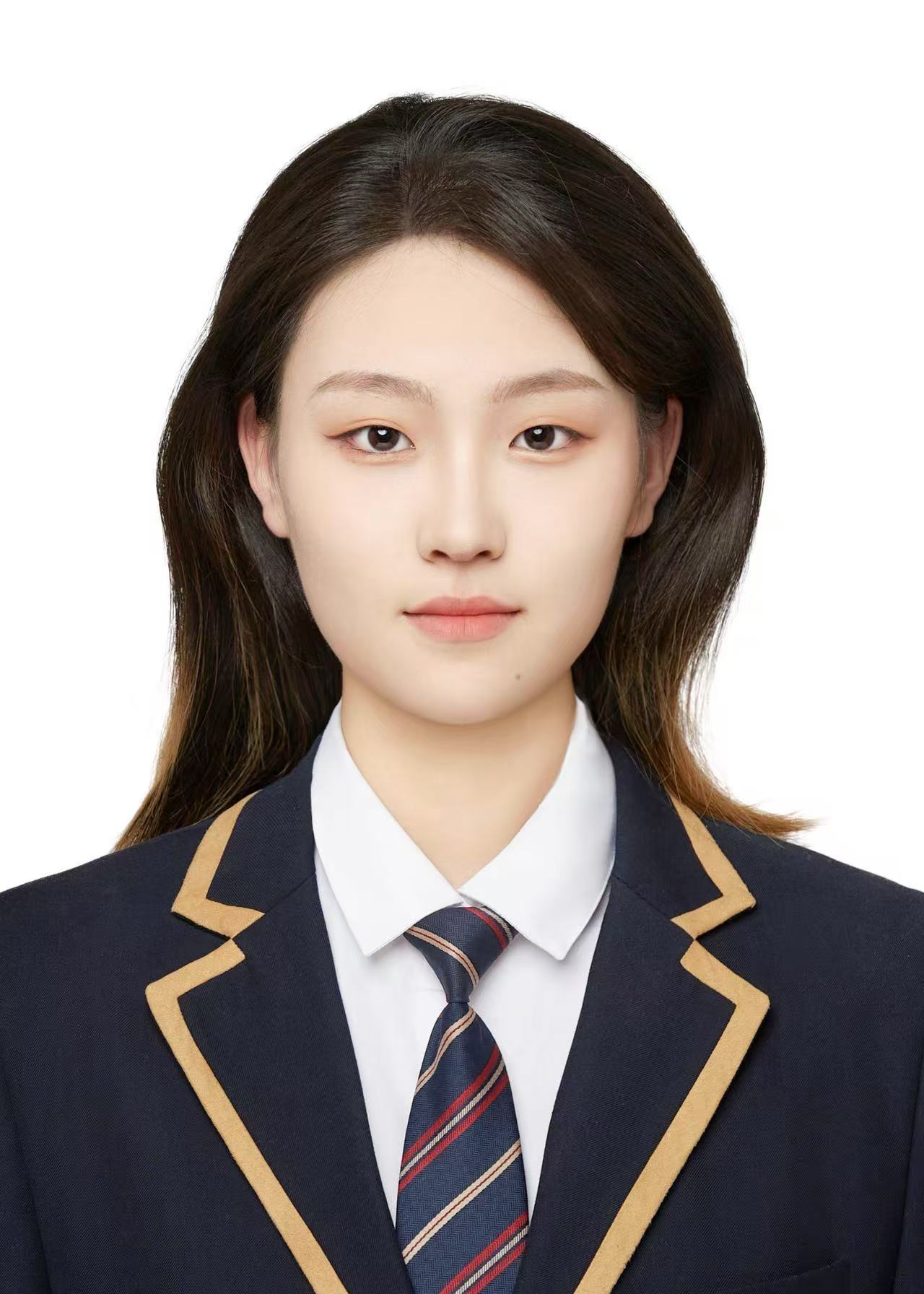}}] 
{Shaoxin Wu} received her bachelor's degree from the School of Computer Science and Technology at Xidian University, China in 2024. She is currently pursuing her master's degree at the same institution. Her research focuses on multimodal learning and action recognition in computer vision.
\end{IEEEbiography}
\vskip -2\baselineskip plus -1fil

\begin{IEEEbiography}
[{\includegraphics[width=1in,height=1.25in,clip,keepaspectratio]{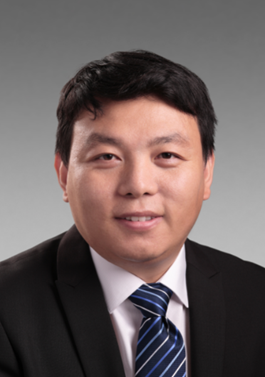}}] 
{Fu Li} is a Professor in the School of Artificial Intelligence at Xidian University, Xi’an, China. He is the head of Xidian-Xilinx Embedded Digital Integrated System Joint Laboratory. He received his B.S. degree in Electronic Engineering from Xidian University in 2004, and Ph.D. degree in Electrical \& Electronic Engineering from the Xidian University in 2010. He has published more than 30 papers international and national journals, and international conferences. His research interests are brain-computer interface, deep learning, small target detection, 3D imaging, embedded deep learning, image and video compression processing, VLSI circuit design, target tracking, neural net- work acceleration and Implementation of intelligent signal processing algorithms (DSP \& FPGA).
\end{IEEEbiography}

\begin{IEEEbiography}
[{\includegraphics[width=1in,height=1.25in,clip,keepaspectratio]{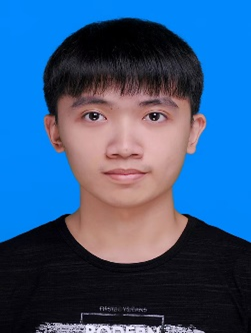}}] 
{Yangtao Zhou} received his B.S. degree in Software Engineering from Xidian University, China, in 2020. He is currently studying as a doctoral student in software engineering at Xidian University, China. His research interests include recommendation systems and data mining.
\end{IEEEbiography}

\vskip -2\baselineskip plus -1fil
\begin{IEEEbiography}
[{\includegraphics[width=1in,height=1.25in,clip,keepaspectratio]{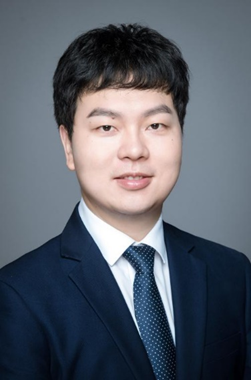}}] 
{Yang Li} received the B.S. degree in electronic information and science technology from School of Physics and Electronics, Shandong Normal University, China, in 2012, the M.S. degree in electronic and communication engineering from School of Electronic Engineering, Xidian University, China, in 2015, the Ph.D. degree from the School of Information Science and Engineering, Southeast University, China, in 2020. He was also a visiting student at University of Wollongong from August 2018 to August 2019. His researches focus on affective computing, pattern recognition and computer vision.
\end{IEEEbiography}

\end{document}